\documentclass[10pt,twocolumn,letterpaper]{article}

\usepackage{cvpr}              %

\usepackage{graphicx}
\usepackage{amsmath}
\usepackage{amssymb}
\usepackage{booktabs}

\usepackage{enumitem}

\usepackage{times}
\usepackage{epsfig}
\usepackage{graphicx}
\usepackage{amsmath}
\usepackage{amssymb}

\usepackage{subcaption} %
\usepackage{booktabs}    %
\usepackage[font=small]{caption}
\usepackage{multirow}
\usepackage{multicol}
\usepackage{arydshln}
\usepackage{capt-of}
\usepackage{algorithm}
\usepackage{listings}
\usepackage{tabularx}
\usepackage{xspace}
\usepackage[table]{xcolor}
\usepackage{colortbl}
\usepackage{cuted}
\usepackage{adjustbox}
\usepackage[pagebackref,breaklinks,colorlinks]{hyperref}

\usepackage[capitalize]{cleveref}
\crefname{section}{Sec.}{Secs.}
\Crefname{section}{Section}{Sections}
\Crefname{table}{Table}{Tables}
\crefname{table}{Tab.}{Tabs.}

\newcommand{\fig}[1]{Figure~\ref{#1}}

\newcommand{\tbl}[1]{Table~\ref{#1}}

\newcommand{\mypar}[1]{\vspace{-5mm}\paragraph{#1}}

\definecolor{lightyellow}{RGB}{255,255,170}

\usepackage{etoolbox}
\makeatletter
\AfterEndEnvironment{algorithm}{\let\@algcomment\relax}
\AtEndEnvironment{algorithm}{\kern2pt\hrule\relax\vskip3pt\@algcomment}
\let\@algcomment\relax
\newcommand\algcomment[1]{\def\@algcomment{\footnotesize#1}}
\renewcommand\fs@ruled{\def\@fs@cfont{\bfseries}\let\@fs@capt\floatc@ruled
  \def\@fs@pre{\hrule height.8pt depth0pt \kern2pt}%
  \def\@fs@post{}%
  \def\@fs@mid{\kern2pt\hrule\kern2pt}%
  \let\@fs@iftopcapt\iftrue}
\makeatother

\newcommand{\supparxiv}[2]{#1}
\def\nondetection{nondetection\xspace}

\newcommand{\projecturl}{https://rrrrrguo.github.io/ganmouflage}

\makeatletter
\DeclareRobustCommand\onedot{\futurelet\@let@token\@onedot}
\def\@onedot{\ifx\@let@token.\else.\null\fi\xspace}
\def\eg{\emph{e.g}\onedot} 
\def\ie{\emph{i.e}\onedot}

\def\etal{\emph{et al}\onedot}
\makeatother

\begin{document}

\title{GANmouflage: 3D Object Nondetection with Texture Fields} %

\author{Rui Guo\textsuperscript{1,3}\thanks{Work done while at University of Michigan}
\and Jasmine Collins\textsuperscript{2}
\and Oscar de Lima\textsuperscript{1}
\and Andrew Owens\textsuperscript{1}
\and
\\ \vspace{-7mm} \and University of Michigan\textsuperscript{1} \vspace{2mm}\\  %
\and UC Berkeley\textsuperscript{2} \and XMotors.ai\textsuperscript{3}~~~~}

\maketitle

\begin{strip}
\vspace{-22mm}
    \centering
    \centering
    \raggedright
    \includegraphics[width=\textwidth]{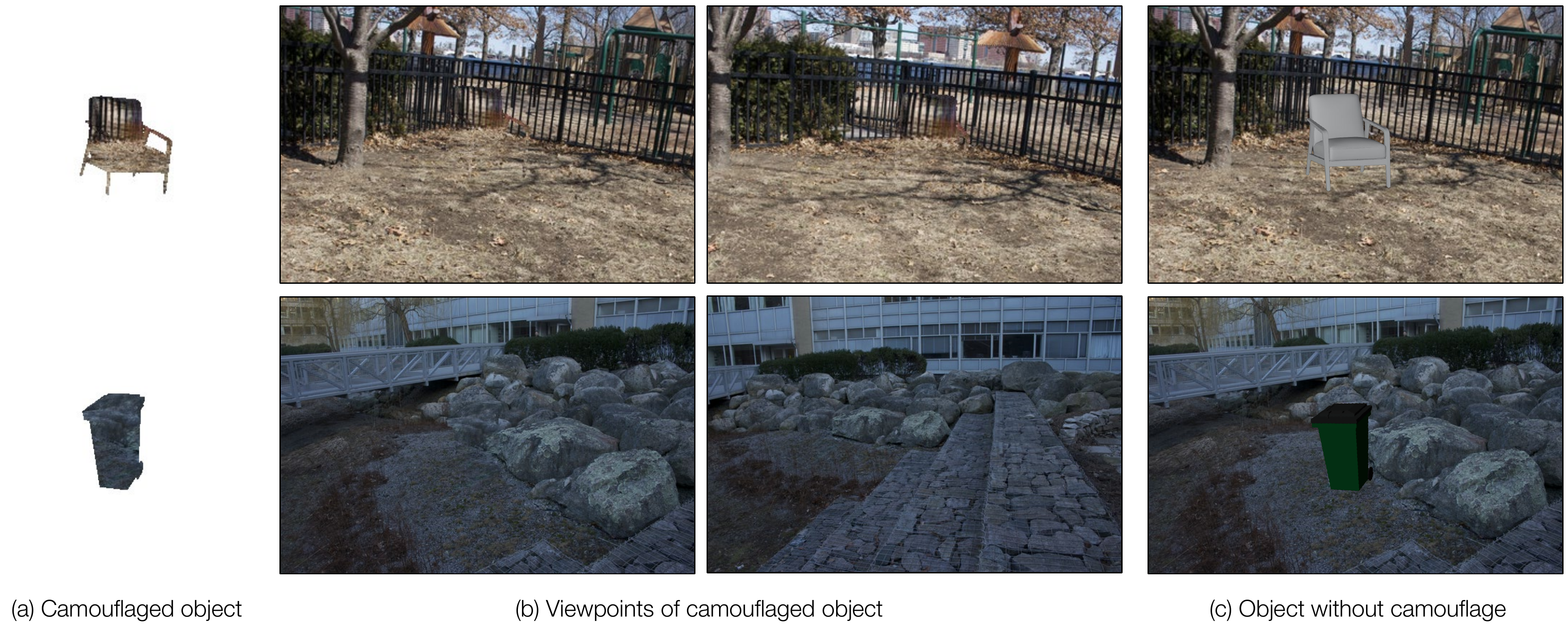}
    \captionof{figure}{
    We learn to camouflage 3D objects within scenes. Given an object's shape, position, and a distribution of possible viewpoints it will be seen from, we estimate a texture field that will conceal it. We show example outputs from our model, with two viewpoints for each camouflaged object. Please see the videos on our \href{\projecturl}{webpage} for more examples.
    }    \vspace{-1mm}
\label{fig:teaser}
\end{strip}

\begin{abstract}
\vspace{-2mm}
We propose a method that learns to camouflage 3D objects within scenes. Given an object's shape and a distribution of viewpoints from which it will be seen, we estimate a texture that will make it difficult to detect. Successfully solving this task requires a model that can accurately reproduce textures from the scene, while simultaneously dealing with the highly conflicting constraints imposed by each viewpoint. We address these challenges with a model based on texture fields and adversarial learning. 
Our model learns to camouflage a variety of object shapes from randomly sampled locations and viewpoints within the input scene, and is the first to address the problem of hiding complex object shapes. Using a human visual search study, we find that our estimated textures conceal objects significantly better than previous methods.

\end{abstract}

\section{Introduction}
\vspace{-2mm}

Using fur, feathers, spots, and stripes, camouflaged animals show a
remarkable ability to stay hidden within their environment. 
These capabilities developed as part of an evolutionary arms race, with
advances in camouflage leading to advances in visual perception, and
vice versa.

Inspired by these challenges, previous work~\cite{owens2014camouflaging} proposed the {\em object nondetection} problem: to create an appearance for an object that makes it undetectable.  Given an object's shape and a sample of photos from a scene, the goal is to produce a texture that hides the object from every viewpoint that it is likely to be observed from. 
This problem  has applications in hiding unsightly objects, such as utility boxes~\cite{callaghanwebsite}, solar panels~\cite{mit2017sistine,wired2016tesla}, 
and radio towers, and in concealing objects from humans or animals, such as surveillance cameras and hunting platforms. Moreover, since camouflage models must ultimately thwart highly effective visual systems, they may provide a better scientific understanding of the cues that these systems use. Animal camouflage, for instance, has developed strategies for avoiding perceptual grouping and boundary detection cues~\cite{talas2019camogan,merilaita2017camouflage}. A successful learning-based camouflage system, likewise, must gain an understanding of these cues in order to thwart them.

Previous object \nondetection methods are based on nonparametric
texture synthesis. %
Although these methods have shown
success in hiding cube-shaped objects, they can only directly ``copy-and-paste" pixels that are directly occluded by the object, making it challenging to deal with complex backgrounds and non-planar geometry. 
While learning-based methods have the potential to address these shortcomings, they face a number of challenges. Since even tiny imperfections in synthesized textures can expose a hidden object, the method must also be capable of reproducing real-world textures with high fidelity. There is also no single texture that can perfectly conceal an object from all viewpoints at once. %
Choosing an effective camouflage requires 3D reasoning, and making trade-offs between different solutions. This is in contrast to the related problem of image inpainting, which can be posed straightforwardly as estimating masked image regions in large, unlabeled photo collections~\cite{pathak2016context}, and which lack the ability to deal with multi-view constraints.

\begin{figure*}[t]
    \vspace{-5mm}
    \centering
    \includegraphics[width=0.98\textwidth]{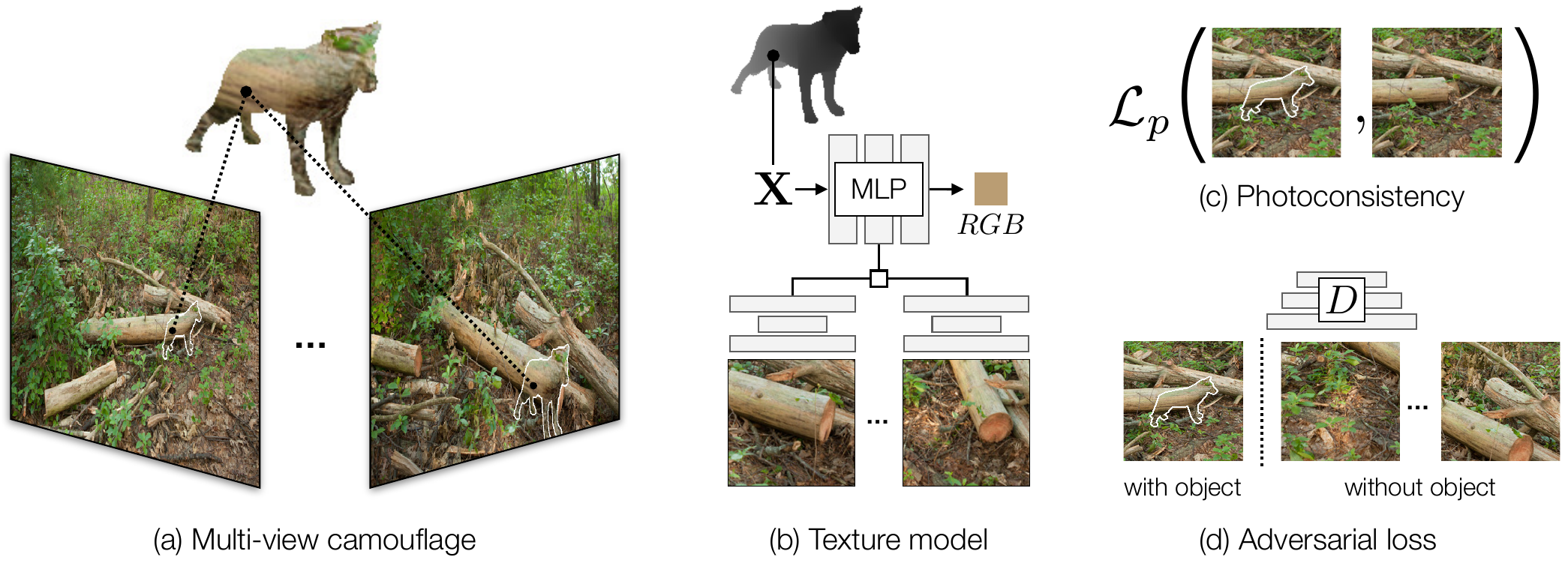}
    \caption{\small \textbf{Camouflage model.}
    (a) Our model creates a texture for a 3D object that conceals it from multiple  viewpoints. (b) We generate a texture field that maps 3D points to colors. The network is conditioned on pixel-aligned features from training images. We train the model to create a texture that is (c) photoconsistent with the input views, as measured using a perceptual loss, and (d) difficult for a discriminator to distinguish from random background patches. For clarity, we show the camouflaged object's boundaries. }
    \vspace{-5mm}
    
    \label{fig:model_pipeline}
\end{figure*}

We propose a model based on neural texture fields~\cite{oechsle2019texture,saito2019pifu,portenier2020gramgan,henzler2020neuraltexture} and adversarial training that addresses these challenges (\fig{fig:model_pipeline}).  The proposed architecture and learning procedure allow the model to exploit multi-view geometry, reproduce a scene's textures with high fidelity, and satisfy the highly conflicting constraints provided by the input images. During training, our model learns to conceal a variety of object shapes from randomly chosen 3D positions within a scene. It uses a conditional generative adversarial network (GAN) to learn to produce textures that are difficult to detect using pixel-aligned representations~\cite{yu2021pixelnerf} with hypercolumns~\cite{hariharan2015hypercolumns} to provide information from each view.

Through automated evaluation metrics and human perceptual studies, we find that our method significantly
outperforms the previous state-of-the-art in hiding cuboid objects. We also
demonstrate our method's flexibility by using it to camouflage a diverse set of complex shapes. These shapes introduce unique challenges, as each viewpoint observes a different set of points on the object surface. Finally, we show through ablations that the design of our texture model leads to significantly better results.

\vspace{-2mm}
\section{Related Work}

\vspace{4mm}
\mypar{Computational camouflage } We take inspiration from early work by Reynolds \cite{reynolds2011interactive} that formulated
 camouflage as part of an artificial life simulation, following Sims~\cite{sims1994evolving} and Dawkins \cite{dawkins1996blind}. In that work, a human ``predator" interactively detects visual ``prey" patterns that are generated using a genetic algorithm. While our model is also trained adversarially, we do so using a GAN, rather than with a human-in-the-loop. %
 Later, Owens~\etal~\cite{owens2014camouflaging} proposed the problem of hiding a cuboid object at a specific location from multiple 3D viewpoints, and solved it using nonparametric texture synthesis. In contrast, our model learns through adversarial training to hide both cuboid and more complex objects. %
Bi~\etal~\cite{bi2017patch} proposed a patch-based synthesis method that they applied to the multi-view camouflage problem, and extended the method to spheres. However, this work was very preliminary: they only provide a qualitative result on a single scene (with no quantitative evaluation).  Other work inserts difficult-to-see patterns into other images~\cite{chu2010camouflage,zhang2020deep}.

\mypar{Animal camouflage.} Perhaps the most well-known camouflage strategy is {\em background matching}, whereby animals take on textures that blend into the background. However,  animals also use a number of other strategies to conceal themselves, such as by masquerading as other objects~\cite{stevens2011animal}, and using {\em disruptive coloration} to elude segmentation cues and to hide conspicuous body parts, such as eyes~\cite{cott1940adaptive}. The object \nondetection problem is motivated by animals that can dynamically change their appearance to match their surroundings, such as the octopus\footnote{For a striking demonstration, see this
video from Roger Hanlon: \url{https://www.youtube.com/watch?v=JSq8nghQZqA}}~\cite{hanlon2007cephalopod}.  Researchers have also begun using computational models to study animal camouflage. Troscianko~\etal~\cite{troscianko2017relative} used a genetic algorithm to camouflage synthetic bird eggs, and asked human subjects to detect them. Talas~\etal~\cite{talas2019camogan} used a GAN to camouflage simple triangle-shaped representations of moths that were placed at random locations on synthetic tree bark. In both cases, the animal models are simplified and 2D, whereas our approach can handle complex 3D shapes.

\mypar{Camouflaged object detection.}
Recent work has sought to detect camouflaged objects using object detectors~\cite{fan2020camouflaged,le2021camouflaged,yan2021mirrornet} and motion cues~\cite{lamdouar2020betrayed,charig2021self}. The focus of our work is generating camouflaged objects, rather than detecting them.

\mypar{Adversarial examples.} The object nondetection problem is related to adversarial examples~\cite{szegedy2013intriguing,goodfellow2014explaining,brown2017adversarial}, in that both problems involve deceiving a visual system (\eg, by concealing an object or making it appear to be from a different class). Other work has generalized these examples to multiple viewpoints~\cite{athalye2018synthesizing}. In contrast, the goal of the nondetection problem is to make objects that are concealed from a human visual system, rather than fool a classifier.

\mypar{Texture fields.} %
We take inspiration from recent work that uses implicit representations of functions to model the surface texture of objects~\cite{oechsle2019texture,henzler2020neuraltexture,portenier2020gramgan,saito2019pifu}. Oechsle~\etal~\cite{oechsle2019texture} learned to texture a given object using an implicit function, with image and shape encoders, and Saito~\etal~\cite{saito2019pifu} learned a pixel-aligned implicit function for clothed humans. There are three key differences between our work and these methods. First, these methods aim to reconstruct textures from given images while our model predicts a texture that can conceal an object. Second, our model is conditioned on a 3D input scene with projective structure, rather than a set of images. Finally, the constraints provided by our images are mutually incompatible: there is no single way to texture a 3D object that satisfies all of the images.
Other work has used implicit functions to represent 3D scenes for view synthesis~\cite{sitzmann2019scene,mildenhall2020nerf,yu2021pixelnerf,chen2021mvsnerf}. %
Sitzmann~\etal~\cite{sitzmann2019scene} proposed an implicit 3D scene representation. Mildenhall~\etal~\cite{mildenhall2020nerf} proposed view-dependent neural radiance fields (NeRF). Recent work created image-conditional NeRFs~\cite{yu2021pixelnerf,chen2021mvsnerf}. Like our method, they use networks with skip connections that exploit the projective geometry of the scene. However, their learned radiance field does not ensure multi-view consistency in color, since colors are conditioned on viewing directions of novel views.

\mypar{Inpainting and texture synthesis. } 
The camouflage problem is related to image inpainting \cite{efros1999texture,bertalmio2000image,hays2007scene,barnes2009patchmatch,pathak2016context,yu2018generative}, in that both tasks involve creating a texture that matches a surrounding region. However, in contrast to the inpainting problem, there is no single solution that can completely satisfy the constraints provided by all of the images, and thus the task cannot be straightforwardly posed as a self-supervised data recovery problem~\cite{pathak2016context}. Our work is also related to image-based texture synthesis~\cite{efros1999texture,barnes2009patchmatch,gatys2016image} and 3D texture synthesis~\cite{oechsle2019texture,henzler2020neuraltexture,portenier2020gramgan}. Since these techniques fill a hole in a single image, and cannot obtain geometrically-consistent constraints from multiple images, they {\em cannot be applied to our method without major modifications}. Nevertheless, we include an inpainting-based baseline in our evaluation by combining these methods with previous camouflage approaches.

\section{Learning Multi-View Camouflage}

Our goal is to create a texture for an object that camouflages it from all of the viewpoints that it is likely to be observed from. 
Following the formulation of Owens et al.~\cite{owens2014camouflaging},  our input is a 3D object mesh $\mathcal{S}$ at a fixed location in a scene, a sample of photos  $I_1, I_2, ..., I_N$
from distribution $\mathcal{V}$, and their camera parameters  $\mathbf{K}_j,\mathbf{R}_j,\mathbf{t}_j$. We desire a solution that camouflages the object from $\mathcal{V}$, using this sample. We are also provided with a ground plane $\mathbf{g}$, which the object has been placed on. 

Also following~\cite{owens2014camouflaging}, we consider the camouflage problem separately from the {\em display} problem of creating a real-world object. We assume that the object can be assigned arbitrary textures, and that there is only a single illumination condition. We note that shadows are independent of the object texture, and hence could be incorporated into this problem framework by inserting shadows into images (Sec.~\ref{sec:qualitative}). Moreover, changes in the amount of lighting are likely to affect the object and background in a consistent way, producing a similar camouflage.

\subsection{Texture Representation}%
We create a surface texture for the object that, on average, is difficult to detect when observed from viewpoints randomly sampled from $\mathcal{V}$. As in prior work~\cite{owens2014camouflaging}, we render the object and synthetically insert it into the scene.

Similar to recent work on object texture synthesis~\cite{oechsle2019texture,henzler2020neuraltexture,portenier2020gramgan}, we represent our texture as continuous function in 3D space, using a {\em texture field}:
\vspace{-2pt}
\begin{equation}
    t_\theta : \mathbb{R}^3 \rightarrow \mathbb{R}^3.
    \label{eq:neuraltexture}
\end{equation}
\vspace{-2pt}
This function maps a 3D point to an RGB color, and is parameterized using a multi-layer perceptron (MLP) with weights $\theta$.

We condition our neural texture representation on input images, %
their projection matrices $\mathbf{P}_j=\mathbf{K}_j[\mathbf{R}_j | \mathbf{t}_j]$, and a 3D object shape $\mathcal{S}$. Our goal is to learn a {\em texturing function} that produces a texture field from an input scene:
\vspace{-2pt}
\begin{equation}
    G_{\theta}(\mathbf{x}; \{\mathbf{I}_j\},\{\mathbf{P}_j\},\mathcal{S})
    \label{eq:main_texture}
\end{equation}
\vspace{-2pt}
where $\mathbf{x}$ is a 3D {\em query point} on the object surface.

\vspace{-2pt}
\subsection{Camouflage Texture Model}
\label{sec:detail_structure}

To learn a camouflaged texture field (Eq. \ref{eq:main_texture}), we require a representation for the multi-view scene content, geometry, and texture field. We now describe these components in more detail. Our full model is shown in~\fig{fig:model_pipeline}.

\mypar{Pixel-aligned image representation.} 
In order to successfully hide an object, we need to reproduce the input image textures with high fidelity. 
For a given 3D point $\mathbf{x}_i$ on the object surface and an image $\mathbf{I}_j$, we compute an image feature $\mathbf{z}_i^{(j)}$ as follows.

We first compute convolutional features for $\mathbf{I}_j$ using a U-net~\cite{unet} with a ResNet-18~\cite{he2016deep} backbone at multiple resolutions. %
We extract image features $\mathbf{F}^{(j)}=E(\mathbf{I}_j)$ %
at full, $\frac{1}{4}$, and $\frac{1}{16}$ scales. At each pixel, we concatenate features for each scale together, producing a multiscale hypercolumn representation~\cite{hariharan2015hypercolumns}.

Instead of using a single feature vector to represent an entire input image, as is often done in neural texture models that create a texture from images~\cite{oechsle2019texture,henzler2020neuraltexture}, we exploit the geometric structure of the multi-view camouflage problem. We extract {\em pixel-aligned} features $\mathbf{z}_i^{(j)}$ from each feature map $\mathbf{F}^{(j)}$, following work in neural radiance fields~\cite{yu2021pixelnerf}. We compute the projection of a 3D point $\mathbf{x}_i$ in viewpoint $\mathbf{I}_j$:
\begin{equation}
    \mathbf{u}_i^{(j)}=\pi^{(j)}(\mathbf{x}_i),
    \label{eq:projection}
\end{equation}
where $\pi$ is the projection function from object space to screen space of image $\mathbf{I}_j$. We then use bilinear interpolation to extract the feature vector $\mathbf{z}_i^{(j)}=\mathbf{F}^{(j)}(\mathbf{u}_i^{(j)})$ for each point $i$ in each input image $\mathbf{I}_j$.

\mypar{Perspective encoding.} 
In addition to the image representation, we also condition our texture field on a {\em perspective encoding} that conveys the local geometry of the object surface and the multi-view setting. For each point $\mathbf{x}_i$ and image $\mathbf{I}_j$, we provide the network with the viewing direction %
 $\mathbf{v}_{i}^{(j)}$ and surface normal $\mathbf{n}_{i}^{(j)}$. %
 These can be computed as:  $\mathbf{v}_{i}^{(j)}=\frac{\mathbf{K}_j^{-1}\mathbf{u}_i^{(j)}}{\|\mathbf{K}_j^{-1}\mathbf{u}_i^{(j)}\|_2}$ and $\mathbf{n}_{i}^{(j)}=\mathbf{R}_j\mathbf{n}_{i}
$, where %
$\mathbf{u}_i^{(j)}$ is the point's projection (Eq.~\ref{eq:projection}) in homogeneous coordinates, %
and $\mathbf{n}_{i}$ is the surface normal in object space. To obtain $\mathbf{n}_{i}$, we extract the normal of the face closet to $\mathbf{x}_i$. %

We note that these perspective features come from the images that are used as {\em input} images to the texture field, rather than the camera viewing the texture, \ie in contrast to neural scene representations~\cite{chen2021mvsnerf,mildenhall2020nerf,yu2021pixelnerf}, our textures are not viewpoint-dependent. %

\mypar{Texture field architecture.} 
We use these features to define a texture field, an MLP that maps a 3D coordinate $\mathbf{x}_i$ to a color $\mathbf{c}_i$ (Eq.~\ref{eq:neuraltexture}).  It is conditioned on the set of image features for the $N$ input images $\{\mathbf{z}_i^{(j)}\}$, as well as the sets of perspective features $\{\mathbf{v}_{i}^{(j)}\}$ and $\{\mathbf{n}_{i}^{(j)}\}$: %
\begin{align}
\vspace{-2mm}
    \mathbf{c}_i= T(\gamma(\mathbf{x}_i); \{\mathbf{z}_i^{(j)}\},\{\mathbf{v}_{i}^{(j)}\},\{\mathbf{n}_{i}^{(j)}\})
    \vspace{-0.5mm}
\end{align}
where $\gamma(\cdot)$ is a positional encoding~\cite{mildenhall2020nerf}. For this MLP, we use a similar architecture as Yu~\etal~\cite{yu2021pixelnerf}. The network is composed of several fully connected residual blocks and has two stages. In the first stage, which consists of 3 blocks, the vector from each input view is processed separately with shared weights. %
Mean pooling is then applied to create a unified representations from the views. In the second stage, another 3 residual blocks are used to predict the color for the input {\em query point}. \supparxiv{Please see the supplementary material for more details}{}. %

\mypar{Rendering.} %
To render the object from a given viewpoint, following the strategy of Oechsle~\etal~\cite{oechsle2019texture}, we determine which surface points are visible  %
using the object's depth map, which we compute using PyTorch3D~\cite{ravi2020pytorch3d}.  %
Given a pixel $\mathbf{u}_i$, we estimate a 3D surface point $\mathbf{x}_i$ in object space through inverse projection: $\mathbf{x}_i= d_i\mathbf{R}^T\mathbf{K}^{-1}\mathbf{u}_i-\mathbf{R}^T\mathbf{t}$,
where $d_i$ is the depth of pixel $i$, $\mathbf{K},\mathbf{R},\mathbf{t}$ are the view's camera  parameters, and $\mathbf{u}_i$ is in homogeneous coordinates. %
We estimate the color for all visible points, and render the object by inserting the estimated pixel colors into a background image, $\mathbf{I}$. This results in a new image that contains the camouflaged object, $\hat{\mathbf{I}}$.

\subsection{Learning to Camouflage}
\label{sec:training}

We require our camouflage model to generate textures that are photoconsistent with the input images, and that are not easily detectable by a learned discriminator.  These two criteria lead us to define a loss function consisting of a photoconsistency term and adversarial loss term, which we optimize through a learning regime that learns to camouflage randomly augmented objects from random positions.

\mypar{Photoconsistency.} The photoconsistency loss measures how well the textured object, when projected into the input views, matches the background. We use a perceptual loss, $\mathcal{L}_{photo}$~\cite{gatys2016image,Johnson2016Perceptual} that is computed as the normalized  distance between activations for layers of a VGG-16 network~\cite{Simonyan15vgg}  trained on ImageNet~\cite{russakovsky2015imagenet}:  %
\begin{align}
    \small
    \mathcal{L}_{photo}=\sum_{j \in J} \mathcal{L}_P(\hat{\mathbf{I}}_j, \mathbf{I}_j) = \sum_{j \in J, k\in L}\frac{1}{N_k}\|\phi_k(\hat{\mathbf{I}}_j)-\phi_k(\mathbf{I}_j)\|_1
    \vspace{-1mm}
    \label{eq:l_feat}
\end{align}
where $J$ is the set of view indices, $L$ is the set of layers used in the loss, and $\phi_k$ are the activations of layer $k$, which has total dimension $N_k$. In practice, due to the large image size relative to the object, we use a crop centered around the object, rather than $\mathbf{I_j}$ itself (see \fig{fig:model_pipeline}(c)).

\begin{figure*}[t]
    \centering
    \includegraphics[width=0.95\textwidth]{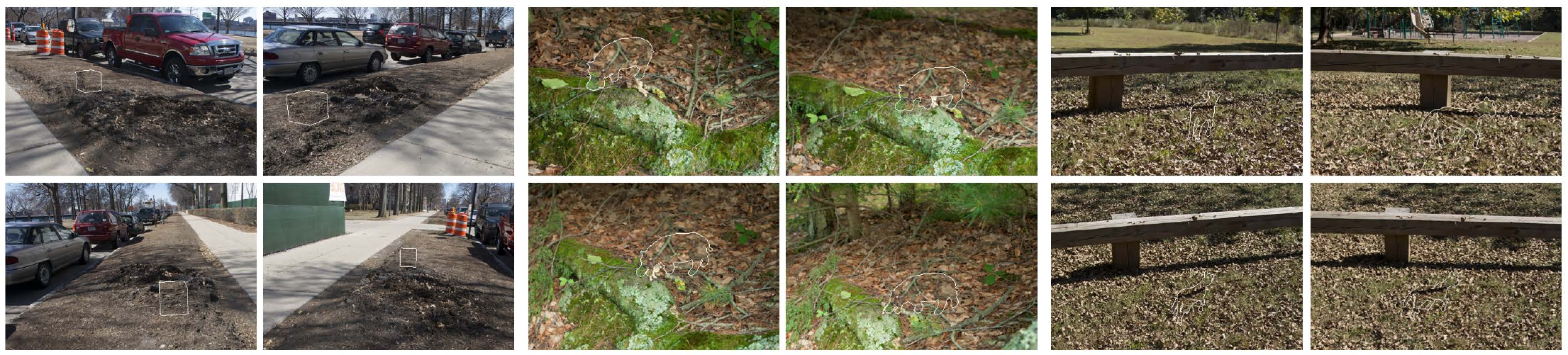}
    \vspace{-2mm}
    \caption{\textbf{Multi-view results}. Multiple object views for selected scenes, camouflaged using our proposed model with four input views. The views shown here were held out and not provided to the network as input during training.}
    \label{fig:more_mutiview}
    \vspace{-3mm}
\end{figure*}

\mypar{Adversarial loss.}
To further improve the quality of generated textures, we also use an adversarial loss. Our model tries to hide the object, while a discriminator attempts to detect it from the scene. %
We randomly select {\em real} image crops $y$ from each background image $\mathbf{I}_j$ and select {\em fake} crops $\hat{y}$ containing the camouflaged object from $\hat{\mathbf{I}}_j$. We use the standard GAN loss as our objective.
To train the discriminator, $D$, we minimize:
\begin{equation}
\vspace{-2mm}
    \mathcal{L}_D= -\mathbb{E}_y[\log D(y)]-\mathbb{E}_{\hat{y}}[\log (1-D(\hat{y}))]
    \vspace{-.5mm}
\end{equation}
where the expectation is taken over patches randomly sampled from a training batch. We implement our discriminator using the fully convolutional architecture of Isola~\etal~\cite{isola2017image}. 
Our texturing function, meanwhile, minimizes:
\begin{equation}
\vspace{-2mm}
    \mathcal{L}_{adv} = -\mathbb{E}_{\hat{y}}[\log D(\hat{y})]
    \vspace{-0.5mm}
\end{equation}

\mypar{Self-supervised multi-view camouflage.} We train our texturing function $G$~(Eq.~\ref{eq:main_texture}), which is fully defined by the image encoder $E$ and the MLP $T$, by minimizing the combined losses:
\begin{align}
    \vspace{-2mm}
    \mathcal{L}_G= \mathcal{L}_{photo} + \lambda_{adv} \mathcal{L}_{adv}
    \label{eq:loss_G}
    \vspace{-1mm}
\end{align}
where $\lambda_{adv}$  controls the importance of the two losses. 

If we were to train the model with only the input object, the discriminator would easily overfit, and our model would fail to obtain a learning signal. Moreover, the resulting texturing model would only be specialized to a single input shape, and may not generalize to others. To address both of these issues, we provide additional supervision to the model by training it to camouflage randomly augmented shapes at random positions, and from random subsets of views.

We sample  object positions on the ground plane $\mathbf{g}$, within a small radius proportional to the size of input object  $\mathcal{S}$.
We uniformly sample a position within the disk to determine the position for the object.  %
In addition to randomly sampled locations, we also randomly scale the object within a range to add more diversity to training data. During training, we randomly select $N_i$ input views and $N_r$ rendering views without replacement from a pool of training images sampled from $\mathcal{V}$. We calculate $\mathcal{L}_{photo}$ on both $N_i$ input views and $N_r$ views while $\mathcal{L}_{adv}$ is calculated on $N_r$ views. %

\vspace{-2mm} 
\section{Results}
We compare our model to previous multi-view camouflage methods using cube shapes, as well as on complex animal and furniture shapes. %

\vspace{-2mm}
\subsection{Dataset}
We base our evaluation on the scene dataset of~\cite{owens2014camouflaging}, placing objects at their predefined locations. Each scene contains 10-25 photos from different locations. During capturing, only background images are captured, with no actual object is placed in the scene. Camera parameters are estimated using structure from motion~\cite{article_sfm_bundler}. To support learning-based methods that take 4 input views, while still having a diverse evaluation set, we use 36 of the 37 scenes (removing one very small 6-view scene). In \cite{owens2014camouflaging}, their methods are only evaluated on cuboid shape, while our method can be adapted to arbitrary shape without any change to the model. To evaluate our method on complex shapes, we generate camouflage textures for a dataset of 49 animal meshes from \cite{Zuffi:CVPR:2017_SMAL}. We also provide a qualitative furniture shape from~\cite{collins2021abo} (Fig.~\ref{fig:teaser}). 

\vspace{-2mm}
\subsection{Implementation Details}
For each scene, we reserve 1-3 images for testing (based on the total number of views in the scene). Following other work in neural textures~\cite{henzler2020neuraltexture}, we train one network per scene. We train our models using the Adam optimizer~\cite{kingma2014adam} with a learning rate of $2\times10^{-4}$ for the texturing function $G$ and $10^{-4}$ for the discriminator $D$. We use $\lambda_{adv}=0.5$ in Eq.~\ref{eq:loss_G}. We resize all images to be $ 384\times 576$ and use square crops of $128\times 128 $ to calculate losses. 

To ensure that our randomly chosen object locations are likely to be clearly visible from the cameras, we randomly sample object positions on the ground plane (the base of the cube in ~\cite{owens2014camouflaging}). We allow these objects to be shifted at most $3 \times$ the cube's length. During training, for each sample, we randomly select $N_i=4$ views as input views and render the object on another $N_r=2$ novel views.  The model is trained with batch size of 8 for approximately 12k iterations.  For evaluation, we place the object at the predefined position from~\cite{owens2014camouflaging} and render it in the reserved test views. 

\begin{figure*}%
    \centering
    \includegraphics[width=0.948\textwidth]{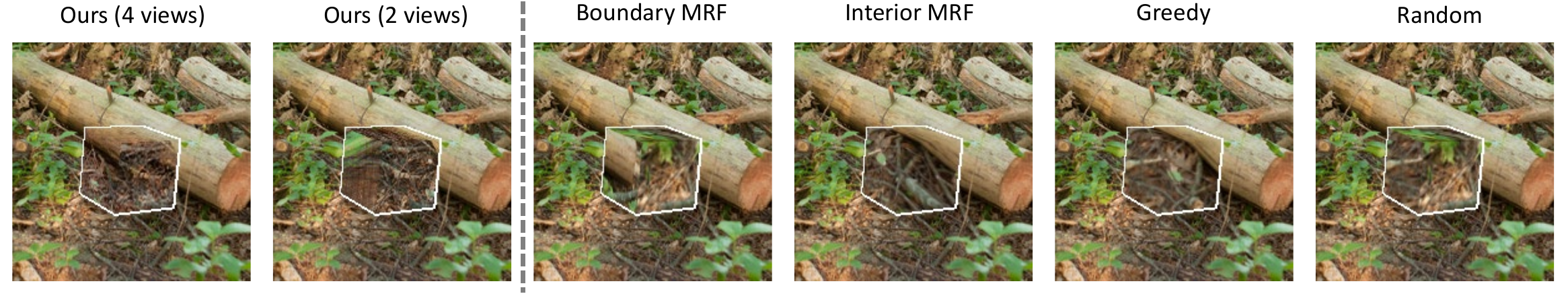}\\
    \centering (a) Qualitative results on cubes\\ \vspace{1mm}
    \includegraphics[width=0.95\textwidth]{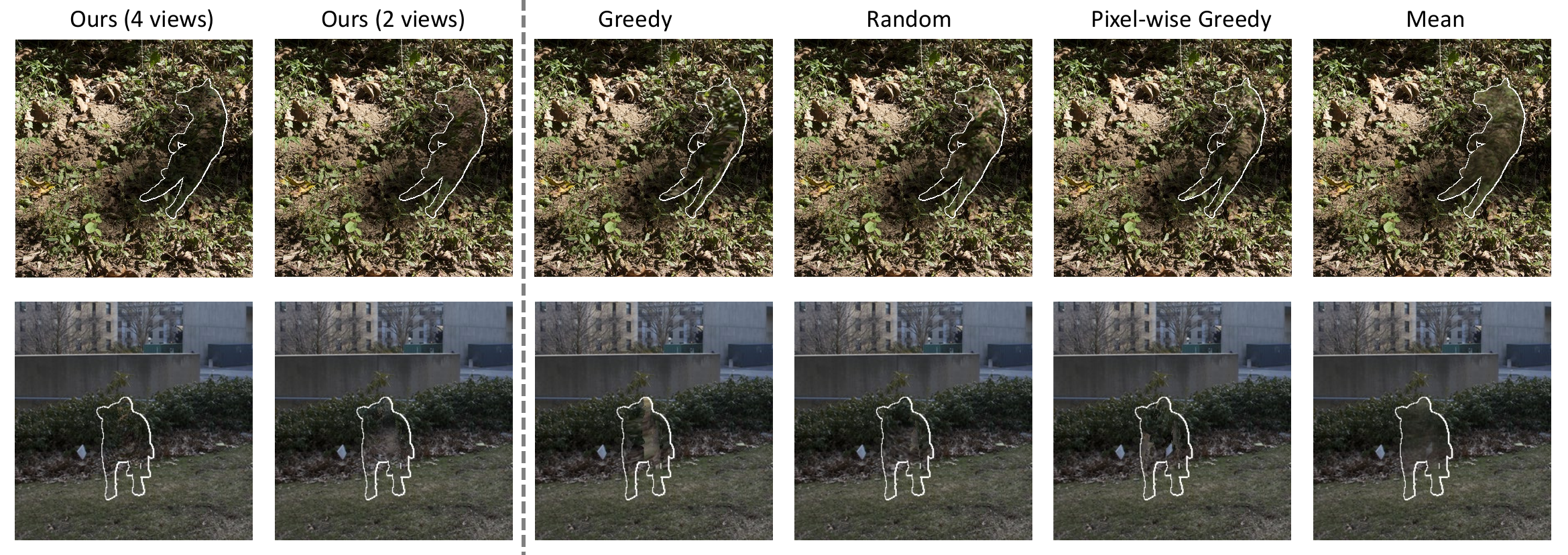}\\
    \centering  (b) Qualitative results on animal shapes
    \vspace{-2mm}
    \caption{\textbf{Comparison between methods for cuboids and complex shapes.} We compare our method with previous approaches for the task of concealing (a) cuboids and (b) animal shapes. Our method produces objects with more coherent texture, with the 4-view model filling in textures that tend to be occluded.} %
    \label{fig:cmp_cube}
    \vspace{-5mm}
\end{figure*}

\subsection{Experimental Settings} 
\subsubsection{Cuboid shapes} 
\vspace{-1mm}
We first evaluate our method using only cuboid shapes to compare with the state-of-the-art methods proposed in Owens et al.~\cite{owens2014camouflaging}. %
We compare our proposed 2-view and 4-view models with the following approaches:

\mypar{Mean.} The color for each 3D point is obtained by projecting it into all the views that observe it and taking the mean color at each pixel.

\mypar{Iterative projection.} These methods exploit the fact that an object can (trivially) be  completely hidden from a single given viewpoint by back-projecting the image onto the object. When this is done, the object is also generally difficult to see from {\em nearby} viewpoints as well. In the \textit{Random} method, the input images are selected in a random order, and each one is projected onto the object, coloring any surface point that has not yet been filled. In \textit{Greedy}, the model samples the photos according to a heuristic that prioritizes viewpoints that observe the object head-on (instead of random sampling). Specifically, the photos are sorted based on the number of object faces that are observed from a direct angle ($>70^\circ$ with the viewing angle). 

\mypar{Example-based texture synthesis.} These methods use Markov Random Fields (MRFs)~\cite{pritch2009shift,agarwala2004interactive,freeman2002example} to perform example-based texture synthesis. These methods simultaneously minimize photoconsistency, as well as smoothness cost that penalizes unusual textures. The {Boundary MRF} model requires nodes within a face to have same labels, while {Interior MRF} does not. %
\vspace{-5mm}
\subsubsection{Complex shapes} %
\vspace{-1mm}
We also evaluated our model on a dataset containing 49 animal meshes~\cite{Zuffi:CVPR:2017_SMAL}. Camouflaging these shapes presents unique challenges. In cuboids, the set of object points that each camera observes is often precisely the same, since each viewpoint sees at most 3 adjacent cube faces (out of 6 total). Therefore, it often suffices for a model to camouflage the most commonly-viewed object points with a single, coherent texture taken from one of the images, putting any conspicuous seams elsewhere on the object. In contrast, when the meshes have more complex geometry, each viewpoint sees a very different set of object points.

Since our model operates on arbitrary shapes, using these shapes requires no changes to the model. We trained our method with the animal shapes and placed the animal object at the same position as in the cube experiments. We adapt the simpler baseline methods of~\cite{owens2014camouflaging} to these shapes, however we note that the MRF-based synthesis methods assume a grid graph structure on each cube face, and hence cannot be adapted to complex shapes without significant changes.

\mypar{Mean.} As with cube experiment, we take the mean color from multiple input views as the simplest baseline.

\mypar{Iterative projection.} We use the same projection order selection strategy as in cube experiment. We determine whether a pixel is visible in the input views by using a ray-triangle intersection test.

\looseness=-1
\mypar{Pixel-wise greedy.} Instead of projecting each input in sequential order, we choose the color for each pixel by selecting color from the input views that has largest view angle.

\begin{table}[]
    \centering
    \resizebox{0.72 \textwidth}{!}{\begin{minipage}{\textwidth}
    \begin{tabular}{lcccc}
      \toprule
    Method & Confusion rate & Avg. time (s)  & Med. time (s) & $n$ \\
      \midrule
    Mean        &16.09$\%\ \pm$ 2.29&4.82 $\pm$ 0.37 &2.95 $\pm$ 0.14 &988\\
    Random      &39.66$\%\ \pm$ 3.02&7.63 $\pm$ 0.50 &4.68 $\pm$ 0.35 &1011\\
    Greedy      &40.32$\%\ \pm$ 2.96&7.94 $\pm$ 0.52 &4.72 $\pm$ 0.36 &1054\\
    Boundary MRF\cite{owens2014camouflaging}&41.29$\%\ \pm$ 2.95&8.50 $\pm$ 0.51 &5.39 $\pm$ 0.40 &1068\\
    Interior MRF\cite{owens2014camouflaging}&44.66$\%\ \pm$ 3.01&8.19 $\pm$ 0.51 &5.19 $\pm$ 0.42 &1048\\
    \cdashline{1-5}
    Ours (2 views)& \textbf{51.58$\%\ \pm$ 2.99}  & \textbf{9.19 $\pm$ 0.51} & \textbf{6.46 $\pm$ 0.42} &1074\\
    Ours (4 views)& \textbf{53.95$\%\ \pm$ 3.05}  & \textbf{9.29 $\pm$ 0.57} & \textbf{6.11 $\pm$ 0.50} &1025\\
    \bottomrule
    \end{tabular}
    \end{minipage}}
    \vspace{-2.5mm}
    \caption{\textbf{Perceptual study results with cubes.} Higher numbers represent a better performance. We report the $95\%$ confidence interval of these metrics.} %
    \label{tab:human_study_cube}
    \vspace{-4.5mm}
\end{table}

\begin{table}
    \centering
     \resizebox{0.49\textwidth}{!}{
     \begin{tabular}{lcccc}
       \toprule
     Method & Confusion rate & Avg. time (s)  & Med. time (s)& $n$ \\
       \midrule
     Mean& 36.46$\%\ \pm$ 2.17 & 6.39 $\pm$ 0.30 & 4.04 $\pm$ 0.17& 1898  \\
     Pixel-wise greedy& 50.43$\%\ \pm$ 2.20 & 7.25 $\pm$ 0.32 & 4.73 $\pm$ 0.20 & 1987\\
     Random& 51.61$\%\ \pm$ 2.29& 7.81 $\pm$ 0.36 & 5.25 $\pm$ 0.36&  1831\\
     Greedy& 52.50$\%\ \pm$ 2.18& 7.69 $\pm$ 0.34  & 5.13 $\pm$ 0.25 & 2017\\
     \cdashline{1-5}
     Ours (4 views)&\textbf{61.93$\%\ \pm$ 2.14} & \textbf{8.06 $\pm$ 0.33}& \textbf{5.66 $\pm$ 0.27}& 1970\\
     \bottomrule
     \end{tabular}
     }
     \vspace{-2mm}
     \caption{\textbf{Perceptual study results on animal shapes.} Higher numbers represent a better performance. We report the $95\%$ confidence interval of these metrics.}
     \vspace{-7mm}
     \label{tab:human_study_animals}
\end{table}

\vspace{-2mm}
\subsection{Perceptual Study}
\vspace{-1mm}
To evaluate the effectiveness of our method, we conduct a perceptual study. We generally follow the setup of~\cite{owens2014camouflaging}, however we ask users to directly click on the camouflaged object~\cite{troscianko2017relative}, without presenting them with a second step to confirm that the object (or isn't) present. This simplified the number of camouflaged objects that subjects see by a factor of two. We recruited 267 and 375 participants from Amazon Mechanical Turk for the perceptual study on cuboid and complex shapes, respectively, and ensured no participant attended both of the perceptual studies. %

Each participant was shown one random image from the reserved images of each scene in a random order. The first 5 images that they were shown were part of a training exercise, and are not included in the final evaluation. We asked participants to search for the camouflaged object in the scene, and to click on it as soon as they found it. The object in the scene was camouflaged by a randomly chosen algorithm, and placed at the predefined position. After clicking on the image, the object outline was shown to the participant. We recorded whether the participant  correctly clicked on the camouflaged object, and how long it took them to click. Each participant had one trial for each image and a maximum of $60s$ to find the camouflaged object.

\mypar{Results on cuboid shapes.} The perceptual study results on cuboid shapes are shown in \tbl{tab:human_study_cube}. We report the confusion rate, average time, and median time measured over different methods. We found that our models {significantly outperform the previous approaches on all metrics}. To test for significance, we followed~\cite{owens2014camouflaging} and used a two-sided $t$-test for the confusion rate and a two-sided Mann-Whitney $U$ test (with a $0.05$ threshold for significance testing). We found that our method outperforms all the baseline methods significantly in the confusion rate metric. Both of our model variations outperform  %
{Interior MRF} ($p < 2 \times 10^{-3}$  and $p<3 \times 10^{-5}$). There was no significant difference between 2 and 4 views ($p=0.28$).
In terms of time-to-click, our method also beats the two MRF-based methods. Compared with {Boundary MRF}, our method requires more time for participants to click the camouflaged object ($p=0.0024$ for 2 views and $p=0.039$ for 4 views). %

\newcolumntype{L}[1]{>{\raggedright\let\newline\\\arraybackslash\hspace{0pt}}m{#1}}
\newcolumntype{C}[1]{>{\centering\let\newline\\\arraybackslash\hspace{0pt}}m{#1}}
\newcolumntype{R}[1]{>{\raggedleft\let\newline\\\arraybackslash\hspace{0pt}}m{#1}}

\begin{figure}
    \begin{center}
    \includegraphics[width=0.98\linewidth]{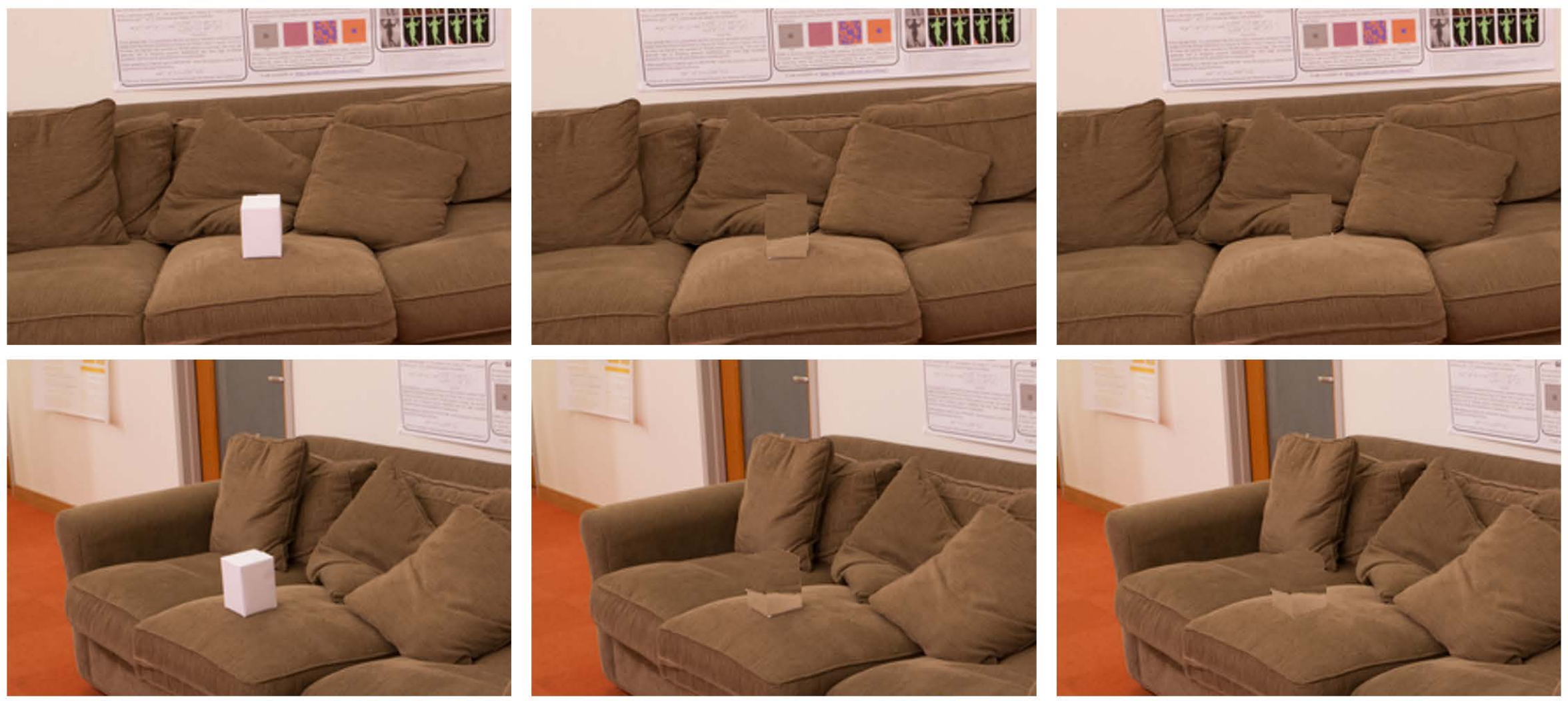}
    \end{center}
    \begin{center} 
    \small
    \vspace{-3mm}
    \begin{tabularx}{\linewidth}{C{0.24\linewidth} C{0.34\linewidth} C{0.24\linewidth}}
    (a) Real cube & (b) With shadow  & (c) No shadow
    \end{tabularx}
    \end{center}
    \vspace{-5mm}
    \caption{{\bf Effect of shadow on generated textures.} We simulate the effect of shadows of the object in an indoor scene, using the reference object (a). Our model generates a texture with a shadow (b) by conditioning on composite images that contain the real shadow (but no real cube). (c) Result without shadow modeling. } 
    \label{fig:shadow}
    \vspace{-6mm}
\end{figure}

\mypar{Results on complex shapes.} The perceptual study results on complex shapes are shown in \tbl{tab:human_study_animals}. We found that our model obtained {\em significantly better} results than previous work on confusion rate. Our model also obtained significantly better results on the time-to-find metric.
We found that in terms of confusion rate, our method with 4 input views is significantly better than the baseline methods, $9.42\%$ better than {Greedy} method and $10.32\%$ better than {Random} method. For time-to-click, our method also performs better than baseline methods compared with {Greedy} and Random.

\vspace{-2mm}
\subsection{Qualitative Results} \label{sec:qualitative}
\vspace{-1mm}
We visualize our generated textures in several selected scenes for both cube shapes and animal shapes in \fig{fig:more_mutiview}. We compare our method qualitatively with baseline methods from \cite{owens2014camouflaging} in \fig{fig:cmp_cube}. We found that our model obtained significantly more coherent textures than other approaches. The 2-view model has a failure case when none of the input views cover an occluded face, while the 4-view model  is able to generally avoid this situation. We provide additional results in the supplement.

\begin{figure}
    \centering
    \includegraphics[width=1.02\linewidth]{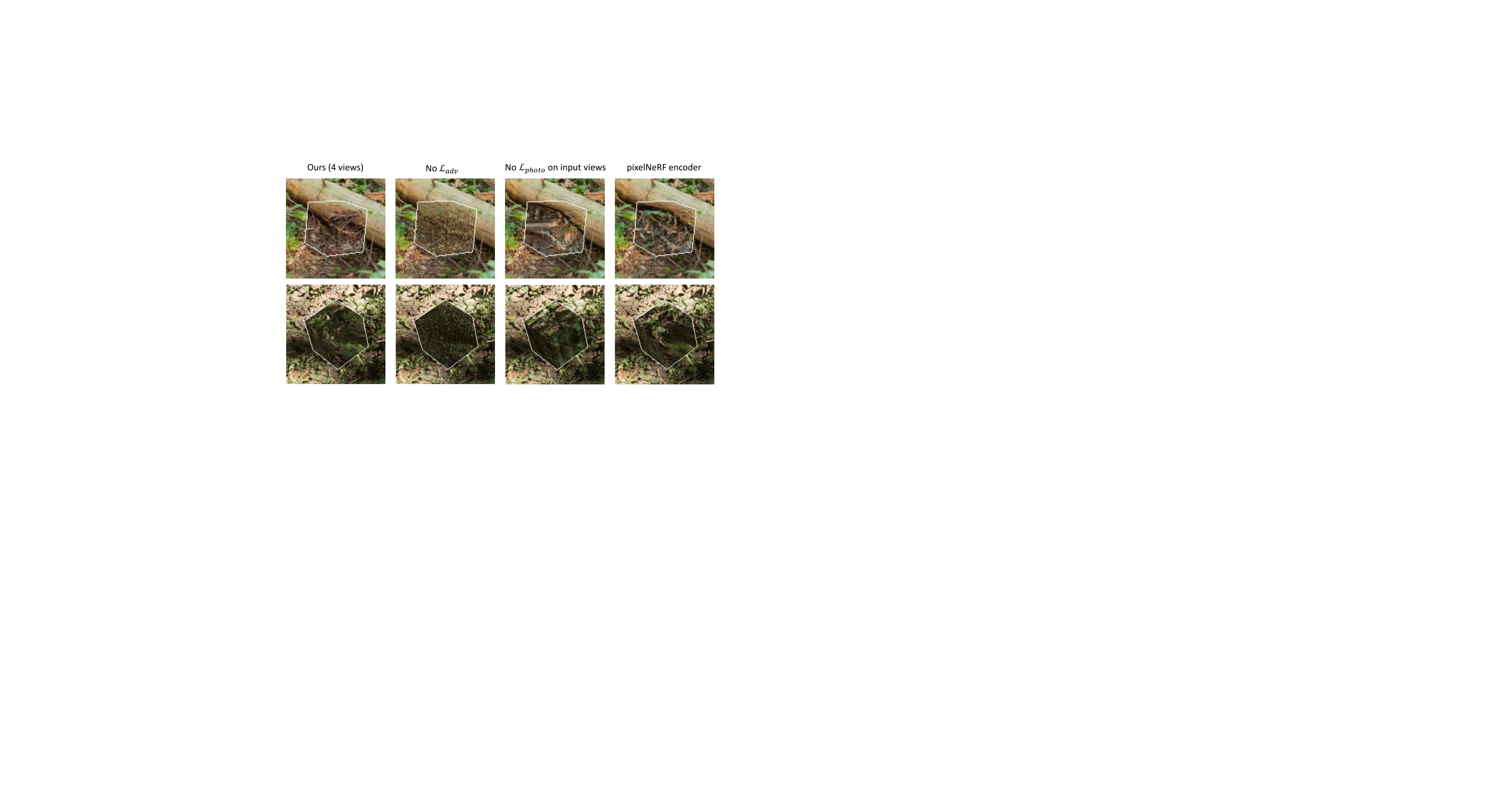}
    \vspace{-6mm}
    \caption{{\bf Ablations.} We show how the choice of different components changes the quality of the camouflage texture.}%
    \label{fig:cmp_ablation}
    \vspace{-5.5mm}
\end{figure}
\mypar{Effects of shadows.} Placing an object in a real scene may create shadows. We ask how these shadows effect our model's solution (\fig{fig:shadow}), exploiting the fact that these shadows are independent of the object's texture and hence function similarly to other image content.  In~\cite{owens2014camouflaging}, photos with (and without) a real cube are taken from the same pose. We manually composite these paired images to produce an image without the real cube but with its real shadow. We then provide these images as conditioning input to our model, such that 
it incorporates the presence of the shadow into its camouflage solution. While our solution incorporates some of the shadowed region, the result is similar. Note that other lighting effects can be modeled as well (\eg, by compensating for known shading on the surface).

\vspace{-2mm}
\subsection{Automated evaluation metrics}
\vspace{-1.5mm}
To help understand our proposed model, we perform an automated evaluation and compare with ablations: %
\vspace{-.25mm}
\begin{itemize}[leftmargin=*,topsep=1pt, noitemsep]
\item {\bf Adversarial loss:} To evaluate the importance of $\mathcal{L}_{adv}$, we set $\lambda_{adv}$ to 0 in Eq.~\ref{eq:loss_G}. We evaluate the model performance with only $\mathcal{L}_{photo}$ used during training.

\item {\bf Photoconsistency:} We evaluate the importance of using all $N_i$ input views in Eq.~\ref{eq:l_feat}. The ablated model has $\mathcal{L}_{photo}$ only calculated on $N_r$ rendering views during training.

\item {\bf Architecture:} We evaluate the importance of our pixel-aligned feature representation. In lieu of this network, we use the feature encoder from pixelNeRF~\cite{yu2021pixelnerf}. %

\item {\bf Inpainting:} Since inpainting methods cannot be directly applied to our task without substantial modifications, we combind several inpainting methods with the {Greedy} model. We selected several recent inpainting methods DeepFillv2~\cite{yu2018deepfillv2}, LaMa~\cite{lama}, LDM~\cite{ldm} to inpaint the object shape in each view, then backproject this texture onto the 3D surface, using the geometry-based ordering from~\cite{owens2014camouflaging}.

\end{itemize}
\vspace{-.5mm}

\mypar{Evaluation metrics.} To evaluate the ablated models, we use LPIPS\cite{zhang2018perceptual} and SIFID metrics~\cite{shaham2019singan}. Since the background portion of the image remains unmodified, we use crops centered at the rendered camouflaged objects.

\mypar{Results.}  Quantitative results are shown in \tbl{tab:ablation} and qualitative results are in \fig{fig:cmp_ablation}. We found that our full 4-view model is the overall best-performing method. In particular, it significantly outperforms the 2-view model, which struggles when the viewpoints do not provide strong coverage from all angles (Fig. \ref{fig:cmp_ablation}). We also found that the adversarial loss significantly improves performance. As can be seen in Fig.~\ref{fig:cmp_ablation}, the model without an adversarial loss fails to choose a coherent solution and instead appears to average all of the input views. The model that uses all views to compute photoconsistency tends to generate more realistic textures, perhaps due to the larger availability of samples. Compared with the pixelNeRF encoder, our model generates textures with higher fidelity, since it receives more detailed feature maps from encoder. We obtain better performance on LPIPS but find that this variation of the model achieves slightly better SIFID. This suggests that the architecture of our pixel-aligned features provides a modest improvement. Finally, we found that we significantly outperformed the inpainting and MRF-based methods.

\begin{table}
\small
    \centering
    \begin{tabularx}{\linewidth}{Xcc}
      \toprule
    Model & LPIPS$\downarrow$ & SIFID$\downarrow$\\
      \midrule
    Boundary MRF~\cite{owens2014camouflaging} &0.1228 & 0.0867\\
    Interior MRF~\cite{owens2014camouflaging}&0.1185 & 0.0782\\
    DeepFill v2~\cite{yu2018deepfillv2} + Projection~\cite{owens2014camouflaging} & 0.1469 & 0.1245 \\
    LaMa~\cite{lama} + Projection~\cite{owens2014camouflaging} & 0.1263 & 0.1006\\
    LDM~\cite{ldm} + Projection~\cite{owens2014camouflaging} & 0.1305 & 0.0976\\
    \cdashline{1-3}
    No $\mathcal{L}_{adv}$&  0.1064& 0.0720 \\
    No $\mathcal{L}_{photo}$ on input views &  0.1131& 0.0856\\
    With pixelNeRF encoder~\cite{yu2021pixelnerf}& 0.1047 & \textbf{0.0712}\\
    Ours (2 views)& 0.1079&  0.0754\\
    Ours (4 views)& \textbf{0.1034} & 0.0714 \\
    \bottomrule
    \end{tabularx}
    \vspace{-3mm}
    \caption{\textbf{Evaluation with automated metrics.} We compare our method to other approaches, and perform ablations.}
    \label{tab:ablation}
    \vspace{-6.5mm}
\end{table}

\vspace{-2.75mm}
\section{Discussion}
\vspace{-1.5mm}
We proposed a method for generating textures to conceal a 3D object within a scene. Our method can handle diverse and complex 3D shapes and significantly outperforms previous work in a perceptual study. We see our work as a step toward developing learning-based camouflage models. Additionally, the animal kingdom has a range of powerful camouflage strategies, such as disruptive coloration and mimicry, that cleverly fool the visual system and may require new learning methods to capture. %

\mypar{Limitations.} As in other camouflage work~\cite{owens2014camouflaging}, we do not address the problem of physically creating the camouflaged object, and therefore do not systematically address practicalities like lighting and occlusion.

\mypar{Ethics.} The research presented in this paper has the potential to contribute to useful applications, particularly to hiding unsightly objects, such as solar panels and utility boxes. However, it also has the potential to be used for negative applications, such as hiding nefarious military equipment and intrusive surveillance cameras.

\mypar{Acknowledgements.} We thank Justin Johnson, Richard Higgins, Karan Desai, Gaurav Kaul, Jitendra Malik, and Derya Akkaynak for the helpful discussions and feedback. This work was supported in part by an NSF GRFP for JC.

{
\bibliographystyle{ieee_fullname}
\bibliography{camo}
}

\appendix
\section*{Appendix}

\section{Implementation details}

\begin{figure}
    \centering
    \includegraphics[width=0.85\linewidth]{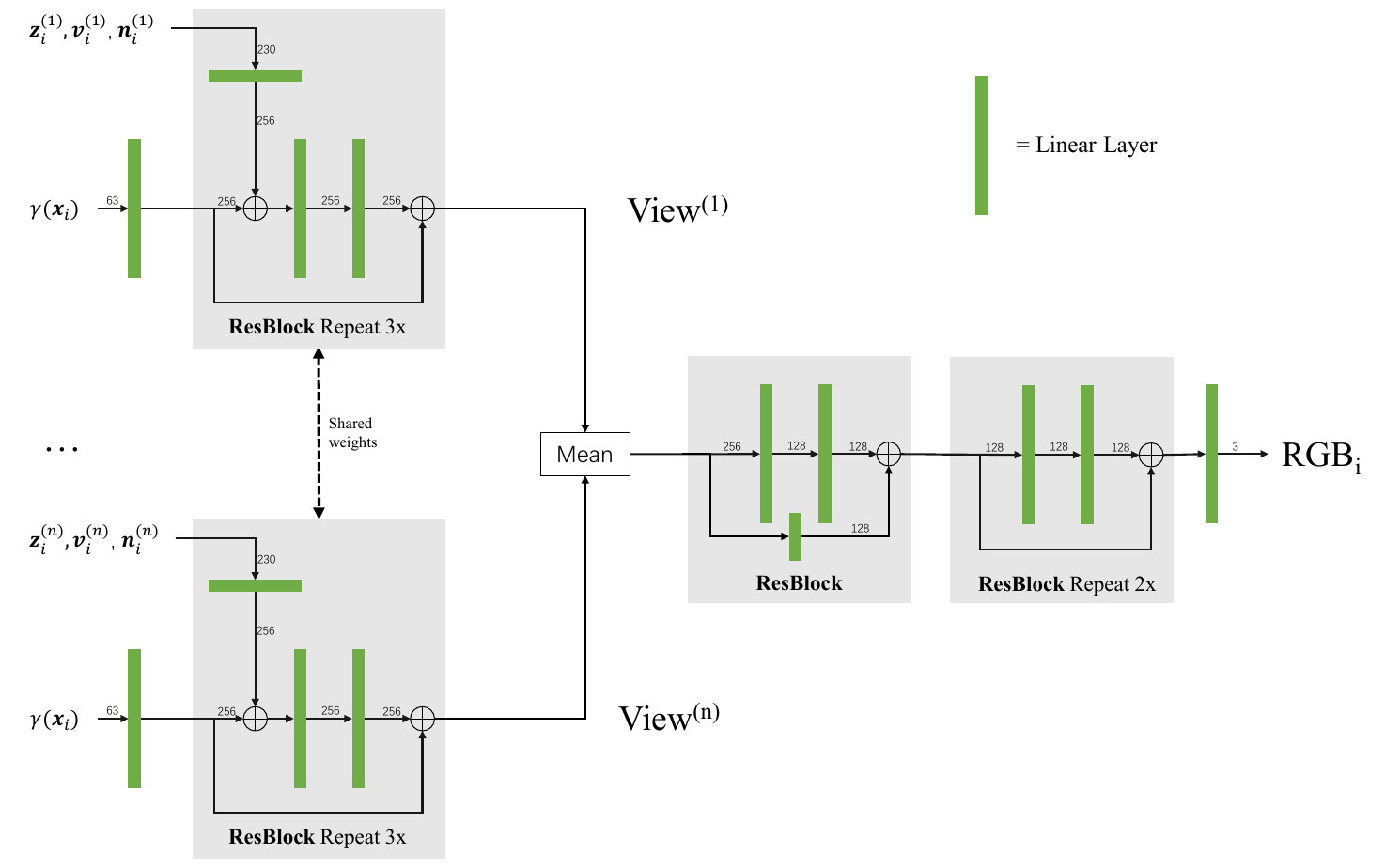}
    \caption{\textbf{Architecture of Multi-Layer Perceptron.} Our model applies a sequence of residual blocks with shared weights to the embedding provided by each viewpoint. We average pool across all views, then predict a color.} 
    \label{fig:mlp_struct}
    \vspace{-5mm}
\end{figure}

\paragraph{Encoder $E$.} In our model, we use a U-net\cite{unet} with ResNet-18\cite{he2016deep} as our feature encoder to extract multi-scale image feature maps from multi-view input images. The U-net has feature maps in 5 different scales in the decoder with shapes:
\begin{enumerate}[topsep=0pt,itemsep=-1ex,partopsep=1ex,parsep=1ex]
    \item $32\times H \times W$
    \item $64\times \frac{H}{2} \times \frac{W}{2}$
    \item $64\times \frac{H}{4} \times \frac{W}{4}$
    \item $128\times \frac{H}{8} \times \frac{W}{8}$
    \item $128\times \frac{H}{16} \times \frac{W}{16}$
\end{enumerate}
 We select 3 level of feature maps($1,\frac{1}{4},\frac{1}{16}$ scales) as the hypercolumn\cite{hariharan2015hypercolumns} features to the lateral multi-layer perceptron $T$, resulting in a total of $224$ channels. We use the implementation of ~\cite{pytorch_unet}.
\mypar{Texture field architecture.}
We base our texture field MLP on~\cite{yu2021pixelnerf}. The detailed structure is shown in \fig{fig:mlp_struct}. The conditional vectors $\{\mathbf{z}_i^{(j)}\},\{\mathbf{v}_i^{(j)}\},\{\mathbf{n}_i^{(j)}\}$ have $230$ channels ($224$ from $\mathbf{z}_i^{(j)}$, $3$ from $\mathbf{v}_i^{(j)}$ and $3$ from $\mathbf{n}_i^{(j)}$) from each input view.  The positional encoding~\cite{mildenhall2020nerf} is computed as:
\begin{align}
    \gamma(\mathbf{x}_i)=\begin{bmatrix}\mathbf{x}_i\\\cos(2^0\mathbf{x}_i)\\\sin(2^0\mathbf{x}_i)\\\vdots\\\cos(2^{L-1}\mathbf{x}_i)\\\sin(2^{L-1}\mathbf{x}_i)\end{bmatrix}.
\end{align}
We set $L=10$, which results in a size of 63 for $\gamma(\mathbf{x}_i)$.

The network is composed of 2 stages. In the first stage, the hidden size is 256 dimensions. We have $N$ separate branches with shared weights for the $N$ input views. After a unified feature representation is generated by mean pooling, the second stage predicts an RGB color for the query point.
\begin{table*}[htbp]
    \centering
    \resizebox{0.9\textwidth}{!}{\begin{minipage}{\textwidth}
    \centering
    \begin{tabular}{lcccc|cc}
      \toprule
    Method & Confusion rate$\uparrow$ & Avg. time$\uparrow$  & Med. time$\uparrow$ & $n$ & LPIPS\cite{zhang2018perceptual}$\downarrow$ & SIFID\cite{shaham2019singan}$\downarrow$ \\
      \midrule
    Mean        &16.09$\%\ \pm$ 2.29&4.82 $\pm$ 0.37 &2.95 $\pm$ 0.14 &988  & 0.1609& 0.1637\\
    Random      &39.66$\%\ \pm$ 3.02&7.63 $\pm$ 0.50 &4.68 $\pm$ 0.35 &1011 & 0.1365& 0.0966\\
    Greedy      &40.32$\%\ \pm$ 2.96&7.94 $\pm$ 0.52 &4.72 $\pm$ 0.36 &1054 & 0.1312& 0.0914\\
    Boundary MRF\cite{owens2014camouflaging}&41.29$\%\ \pm$ 2.95&8.50 $\pm$ 0.51 &5.39 $\pm$ 0.40 &1068 & 0.1228 & 0.0867\\
    Interior MRF\cite{owens2014camouflaging}&44.66$\%\ \pm$ 3.01&8.19 $\pm$ 0.51 &5.19 $\pm$ 0.42 &1048 & 0.1185& 0.0782\\
    \cdashline{1-7}
    Ours (2 views)& \textbf{51.58$\%\ \pm$ 2.99}  & \textbf{9.19 $\pm$ 0.51} & \textbf{6.46 $\pm$ 0.42} &1074 & 0.1079 & 0.0754\\
    Ours (4 views)& \textbf{53.95$\%\ \pm$ 3.05}  & \textbf{9.29 $\pm$ 0.57} & \textbf{6.11 $\pm$ 0.50} &1025 & \textbf{0.1034} & \textbf{0.0714}\\
    \bottomrule
    \end{tabular}
    \end{minipage}}
    \caption{\textbf{Quantitative results with cubes.}} %
    \label{tab:human_study_cube_supp}

\end{table*}

\begin{table*}[htbp]
    \centering
    \resizebox{0.9\textwidth}{!}{\begin{minipage}{\textwidth}
    \centering
    \begin{tabular}{lcccc|cc}
      \toprule
    Method & Confusion rate$\uparrow$ & Avg. time$\uparrow$  & Med. time$\uparrow$ & $n$ & LPIPS\cite{zhang2018perceptual}$\downarrow$ & SIFID\cite{shaham2019singan}$\downarrow$ \\
      \midrule
    Mean& 36.46$\%\ \pm$ 2.17 & 6.39 $\pm$ 0.30 & 4.04 $\pm$ 0.17& 1898 & 0.0883&0.0441 \\
    Pixel-wise greedy& 50.43$\%\ \pm$ 2.20 & 7.25 $\pm$ 0.32 & 4.73 $\pm$ 0.20 & 1987&0.0976&0.0590\\
    Random& 51.61$\%\ \pm$ 2.29& 7.81 $\pm$ 0.36 & 5.25 $\pm$ 0.36&  1831&0.0888&0.0418\\
    Greedy& 52.50$\%\ \pm$ 2.18& 7.69 $\pm$ 0.34  & 5.13 $\pm$ 0.25 & 2017& 0.0881 & 0.0419\\
    \cdashline{1-7}
    Ours (4 views)&\textbf{61.93$\%\ \pm$ 2.14} & \textbf{8.06 $\pm$ 0.33}& \textbf{5.66 $\pm$ 0.27}& 1970 & \textbf{0.0798} & \textbf{0.0350}\\
    \bottomrule
    \end{tabular}
    \end{minipage}}
    \caption{\textbf{Quantitative results with animal shapes.}}
    \label{tab:human_study_animals_supp}
\end{table*}
\mypar{Discriminator $D$.}
For the discriminator, we use the model of~\cite{pix2pix2017}, after replacing the batch normalization layers with instance normalization layers. The network is composed of a sequence of Convolution-InstanceNorm-LeakyReLu blocks. In particular, it has structures of channels of $64,128,256,512$ in each of its blocks. Instance normalization is not applied to the first block.

\section{Additional Results}
\vspace{5mm}
\mypar{Quantitative metrics.}
We show the calculated LPIPS~\cite{zhang2018perceptual} and SIFID~\cite{shaham2019singan} scores on crops of rendered camouflaged objects in \tbl{tab:human_study_cube_supp} (cube) and \tbl{tab:human_study_animals_supp} (animal shapes). Objects are placed at the exactly same place in each scene for each method. We use a square crop centered at the object that has size of $32\lceil d/32\rceil+32$, where $d$ is the maximum dimension of a foreground camouflaged object. We calculate these two metrics on all test views and report the average scores on all scenes.

\begin{figure*}
    \centering
    \includegraphics[width=0.9\textwidth]{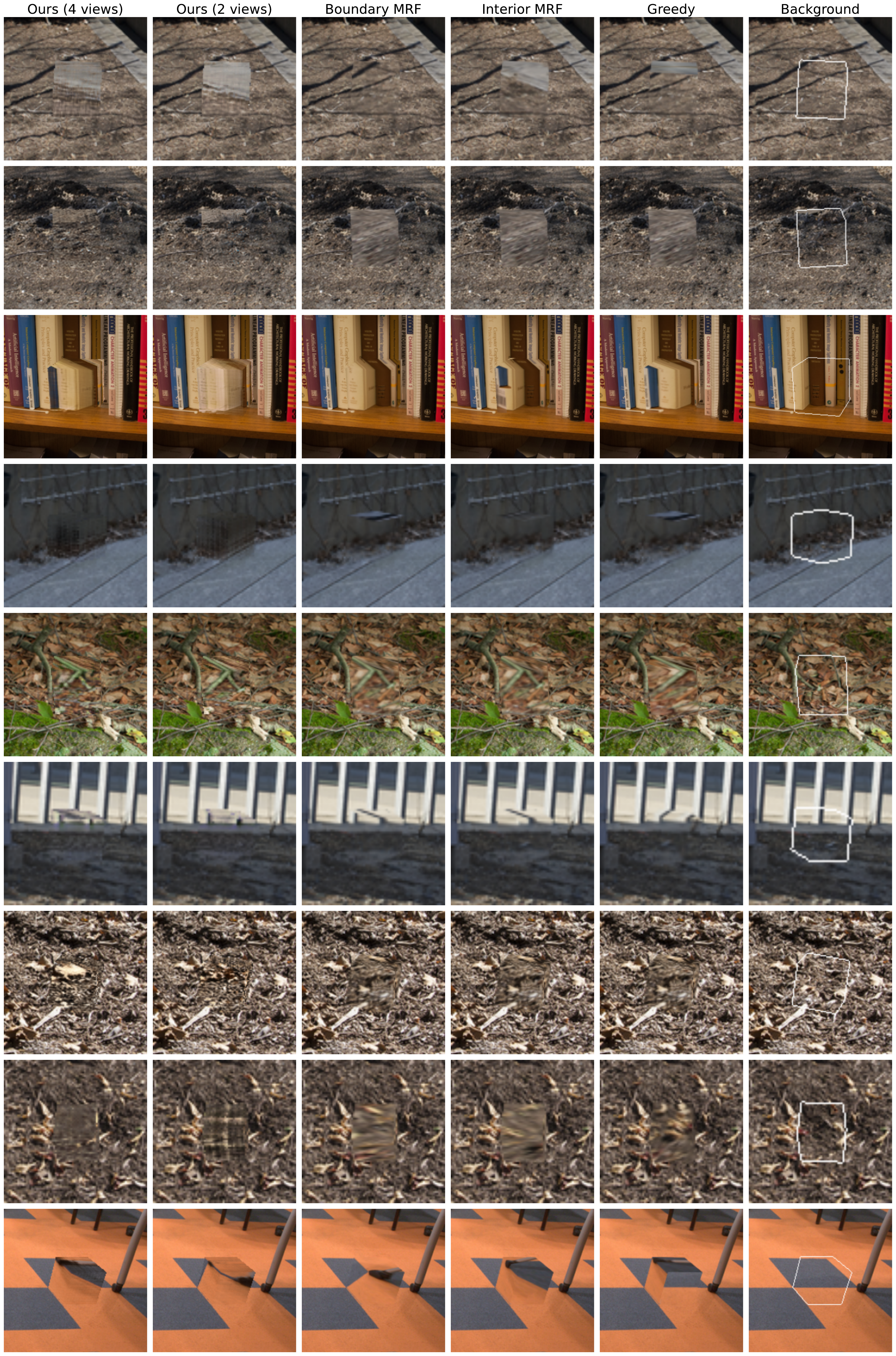}
    \caption{Comparison between methods for camouflaging cubes.}
    \label{fig:vis_cube}
\end{figure*}

\begin{figure*}
    \centering
    \includegraphics[width=0.9\textwidth]{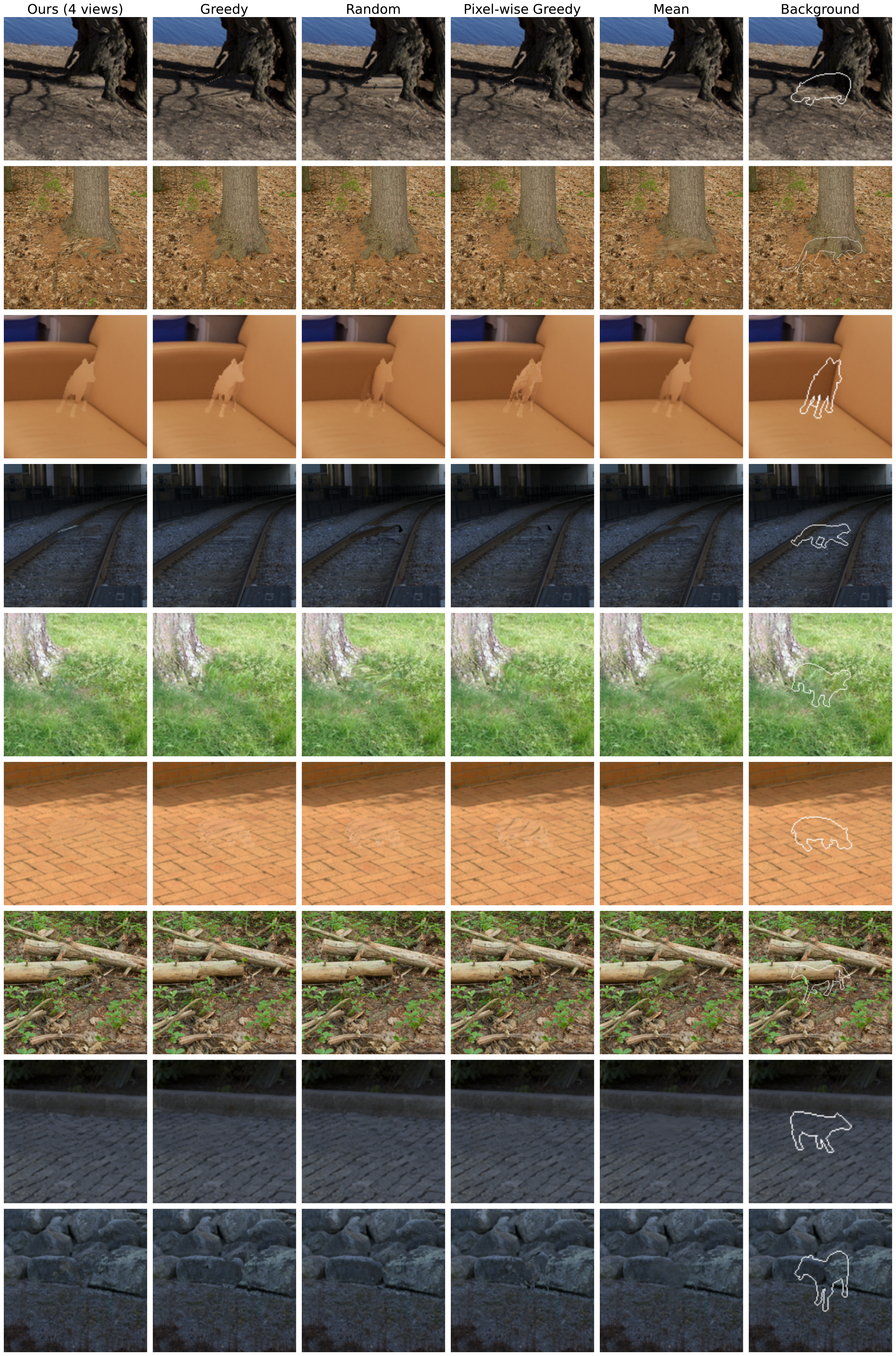}
    \caption{Comparison between methods for camouflaging animals (page 1 of 3).}
    \label{fig:vis_animals1}
\end{figure*}

\begin{figure*}
    \centering
    \includegraphics[width=0.9\textwidth]{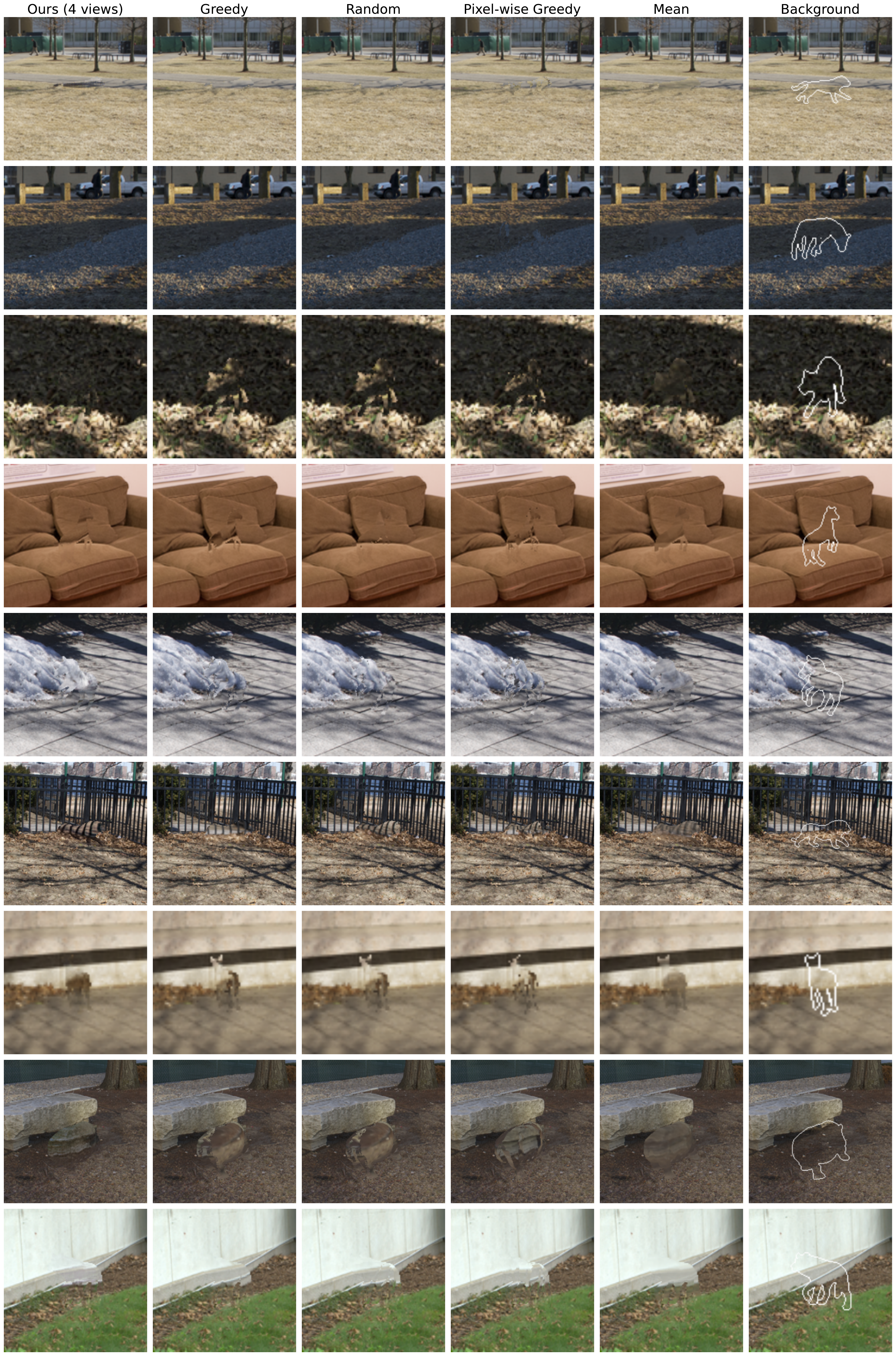}
    \caption{Comparison between methods for camouflaging animals (page 2 of 3)}
    \label{fig:vis_animals2}
\end{figure*}

\begin{figure*}
    \centering
    \includegraphics[width=0.9\textwidth]{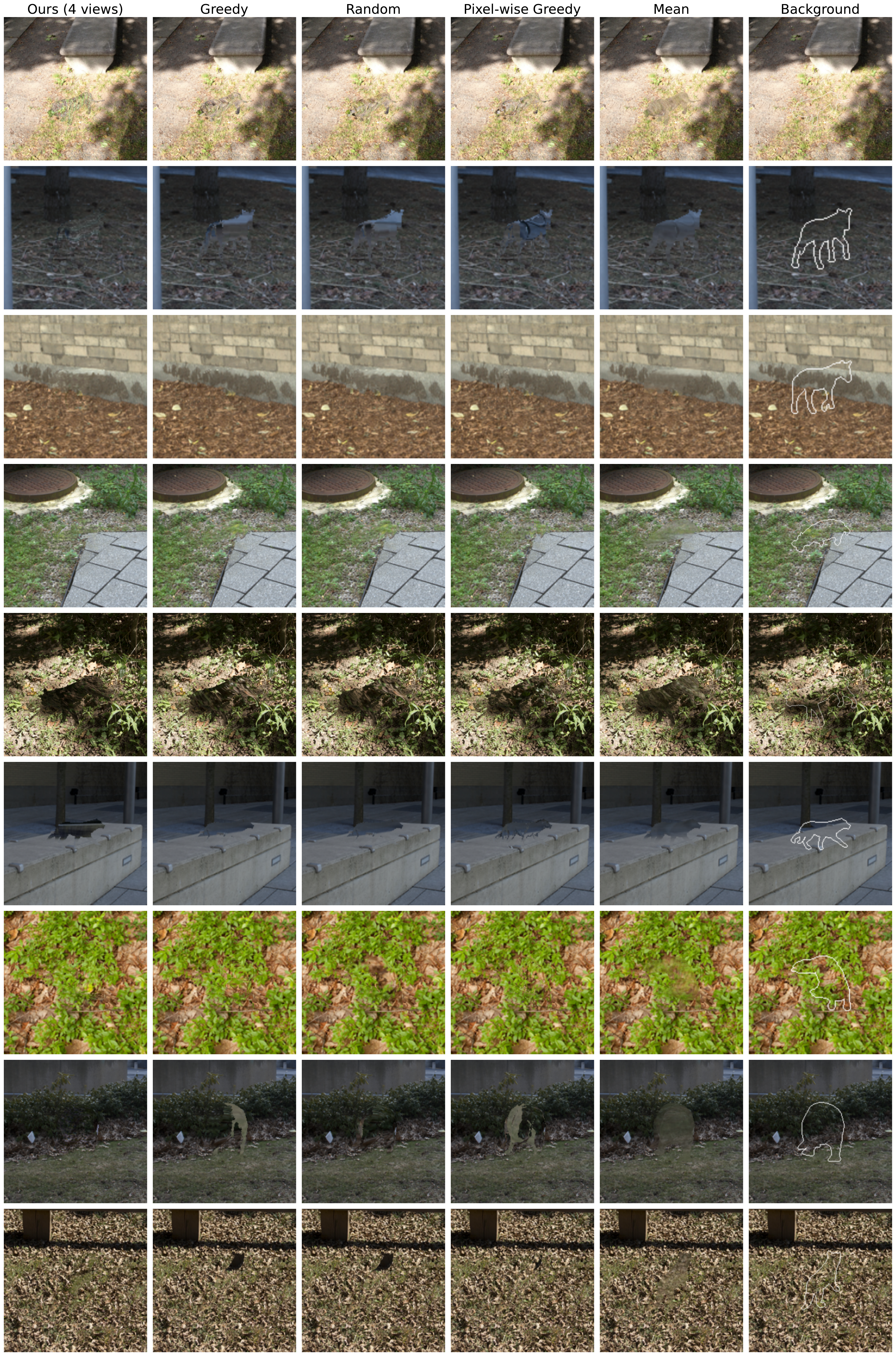}
    \caption{Comparison between methods for camouflaging animals (page 3 of 3)}
    \label{fig:vis_animals3}
\end{figure*}

\mypar{Qualitative Results.}
We show the 36 scenes used in our evaluation, along with the different methods. Cube results are shown in \fig{fig:vis_cube}, and animal results are shown in Figure~\ref{fig:vis_animals1}-\ref{fig:vis_animals3}. The viewpoint and animal shapes are {\bf randomly selected} for these visualizations. We crop the images around the objects to show them more clearly.

\end{document}